\documentclass[a4paper, 10pt, conference]{ieeeconf}      

\IEEEoverridecommandlockouts                              

\usepackage{amsmath,amsfonts}
\usepackage[algo2e,ruled,vlined,linesnumbered,resetcount,norelsize]{algorithm2e}
\usepackage{array}
\usepackage[caption=false,font=normalsize,labelfont=sf,textfont=sf]{subfig}
\usepackage{textcomp}
\usepackage{stfloats}
\usepackage{float}
\usepackage{url}
\usepackage{verbatim}
\usepackage{graphicx}
\usepackage{cite}
\usepackage{mathtools} 
\usepackage{amssymb}  

\usepackage{amsthm}
\usepackage{algpseudocode}
\usepackage[table]{xcolor}
\usepackage{accents}
\usepackage{mathrsfs}
\usepackage{multirow}
\usepackage[normalem]{ulem}

\def\beq{\begin{equation*}}
\def\eeq{\end{equation*}}
\def\bql{\begin{equation}}
\def\eql{\end{equation}}
\def\bqn{\begin{eqnarray*}}
\def\eqn{\end{eqnarray*}}
\def\bnl{\begin{eqnarray}}
\def\enl{\end{eqnarray}}

\usepackage{soul}
\usepackage{booktabs}

\usepackage[T1]{fontenc}
\usepackage[utf8]{inputenc}
\usepackage{siunitx} 
\newcolumntype{T}{S[table-format=3.3, input-symbols={()},
                    table-space-text-post={$^{***}$},
                    table-align-text-post=false]}
\usepackage[colorlinks=true, allcolors=black]{hyperref}
\usepackage{tabularx,booktabs}
\newcolumntype{C}{>{\centering\arraybackslash}X} 
\definecolor{green}{RGB}{11,155,13}
\newcommand{\marginXW}[1]

\usepackage{yaacro}

\begin{acgroupdef}[list=acronimos]
    \acdef{JSG}{Joint State Graph}
\end{acgroupdef}

\title{\LARGE \bf
Manta Ray Inspired Flapping-Wing Blimp
}

\author{Kentaro Nojima-Schmunk$^{1}$, David Turzak$^{1}$, Kevin Kim$^{1}$, Andrew Vu$^{1}$, James Yang$^{2}$\\ Sreeauditya Motukuri$^{3}$, Ningshi Yao$^{2}$, and Daigo Shishika$^{1}$
\thanks{$^{1}$Kentaro Nojima-Schmunk, David Turzak, Kevin Kim, Andrew Vu, and Daigo Shishika are with Mechanical Engineering Department, George Mason University, Fairfax, VA, 22030, USA.}%
\thanks{$^{2}$James Yang and Ningshi Yao are with the Department of Electrical and Computer Engineering, George Mason University, Fairfax, VA, 22030, USA.}%
\thanks{$^{3}$Sreeauditya Motukuri is with the Department of Mechanical Engineering, Virginia Tech, Blacksburg, VA, 24061, USA.}%
}

\begin{document}

\maketitle
\thispagestyle{empty}
\pagestyle{empty}

\begin{abstract}

Lighter-than-air vehicles or blimps, are an evolving platform in robotics with several beneficial properties such as energy efficiency, collision resistance, and ability to work in close proximity to human users. While existing blimp designs have mainly used propeller-based propulsion, we focus our attention to an alternate locomotion method, flapping wings. Specifically, this paper introduces a flapping-wing blimp inspired by manta rays, in contrast to existing research on flapping-wing vehicles that draw inspiration from insects or birds. We present the overall design and control scheme of the blimp as well as the analysis on how the wing performs. 
The effects of wing shape and flapping characteristics on the thrust generation are studied experimentally. We also demonstrate that the flapping-wing blimp has a significant range advantage over propeller-based systems.
\end{abstract}

\section{Introduction}
Lighter-than-air (LTA) vehicles or blimps are an interesting platform for aerial autonomy and have been attracting interest in the robotics community \cite{Motoyama2003, ferdous2019developing}. 
{Some advantages of working with blimps come from the non-rigid airship design providing safety around human users and robustness to collisions~\cite{Monocular_vision,srisamosorn2020indoor}.} 
Another feature of LTA vehicles is the naval relevance; LTA vehicles demonstrate fluid mechanics enabling an aquatic design without dependence on water.
Recent research includes control methods \cite{liu2022deep, ko2007gaussian, tao2021swing}, Human Robot Interactions \cite{yao2019autonomous, hou2019modeling, cho2017autopilot}, and localization \cite{seguin2020deep, lin2023monocular}.
However, the means of locomotion in existing autonomous blimp research is simplified to a combination of propellers that provide sufficient control authority. In this paper we focus on the locomotion aspect and investigate the use of flapping wings as an alternative propelling mechanism.

Focus on the development of flapping-wing robots have mostly come from terrestrial animals. There are plenty of examples inspired by hummingbirds \cite{fei2019flappy, fujii2023hummingbird}, insects \cite{lynch2022autonomous, hollenbeck2012methods, phan2020towards}, and even bats \cite{batmav}. Most of the actuation methods proposed for flapping-wing micro-aerial vehicles (FWMAVs) are not transferable to a larger scale vehicle like a blimp. For our LTA vehicle, we find inspiration from one of the most efficient swimmers, the manta ray \cite{dewey2012relationship, subramanian2021aerodynamic}.

\begin{figure}
    \centering    \includegraphics[width=0.9\linewidth]{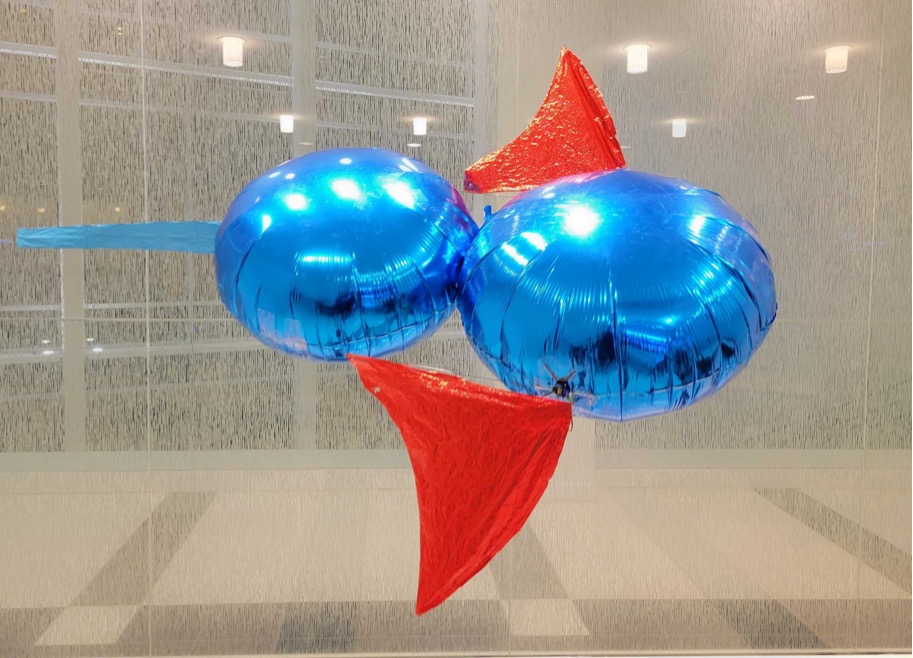}
    \caption{The flapping-wing LTA vehicle: ``Flappy''.}
    \label{fig:Flappy}
    \vspace{-25pt}
\end{figure}

There are existing robots inspired by manta rays, but they are underwater vehicles \cite{chew2015development, zhong2022development}. As a major difference from those studies, one of the challenges in incorporating the flapping wing mechanism for LTA vehicles is the limited payload. The buoyancy achieved by helium in the air is much less than the buoyancy given by air in water. Some sophisticated undulating mechanisms built for underwater robots are realized by a series of strong actuators and mechanical components that a blimp will not be able to carry~\cite{Velox}. 

There are existing flapping-wing blimp robots that vary in size and flapping mechanisms. For example, the Air Ray by Festo \cite{AirRay} and the Cornell ornithoptic blimp \cite{dietl2011dynamic} have large payloads of 1.6~kg and 420~g respectively, while a micro blimp utilizing insect-sized wings \cite{xian2024controlled} and the FlyJelly \cite{yun2023untethered} have small payloads of 2.27~g and 257.9~mg respectively. The smaller payload blimps also utilize complex actuators, such as electromagnetics \cite{xian2024controlled}, electrostatics \cite{yun2023untethered}, and shape memory alloys \cite{barrett2010development} that offer small amounts of thrust. This paper studies a simple minimally viable approach in reproducing efficient flapping motion inspired by manta rays that is comparable to propeller motors.

The shape and structure of the wings play a key role in the behavior of an agent. Flexibility along the manta ray's pectoral fins has been well documented as a reason for their swimming efficiency \cite{fish2016hydrodynamic}. Their fins are rigid near the root with the thicker tissue, then become thinner and more flexible near the tip with cartilage \cite{he2022effects}. We replicate a flexible wing using thin carbon fiber rods.

The two main contributions of this paper are: the design of a low-cost LTA vehicle that moves with manta ray inspired flapping wings and tail; and an analysis of the wing shape and flapping parameters. We also show how our design is more efficient than a propeller-based agent, which results in an increase in range. We believe this platform serves as a base line for the exploration of flapping-wing blimps.
Furthermore, due to its simplicity and low cost (each vehicle can be made with less than 100 USD using off-the-shelf components), the platform is also appropriate for STEM education and outreach.

\section{Vehicle Design}
\label{sec:vehicle_design}
{This section introduces the basic design of the vehicle including the main body and electrical components.}

\subsection{Vehicle Body}
\label{sub:vehicle_body}
Flappy is made of two 91.4 cm diameter ellipsoidal balloons filled with balloon-grade helium. When fully inflated, each balloon supports 68~g of lift, giving total payload of 136~g. The two balloons are attached together with adhesives, and 1.5~mm carbon-fiber rods act as supports as shown in Fig.~\ref{fig:schematic}a. 

\subsection{Actuators}
\label{sub:actuators}
Flappy has two wings and one tail, all actuated by S51-9g plastic gear servos that can move 180$^\circ$. 3D printed servo mounts (Fig.~\ref{fig:schematic}b) are taped to the sides of the front balloon for the wings and one on the back balloon for the tail.  
The servos are wired to an Adafruit HUZZAH32-ESP32 Feather microcontroller. Power is supplied using a 2s 300 mAh LiPo battery. To meet the operable voltage of the Feather and servos, a buck converter down to 5 V is required. The wiring of Flappy is shown in Fig.~\ref{fig:schematic}a. Flappy is controlled wirelessly using a video game controller through Bluetooth Classic connection with the Feather. 

\subsection{Wings and Tail}
\label{sub:wing_tail_material}

\begin{figure}[t]
    \centering
    \includegraphics[width=\linewidth]{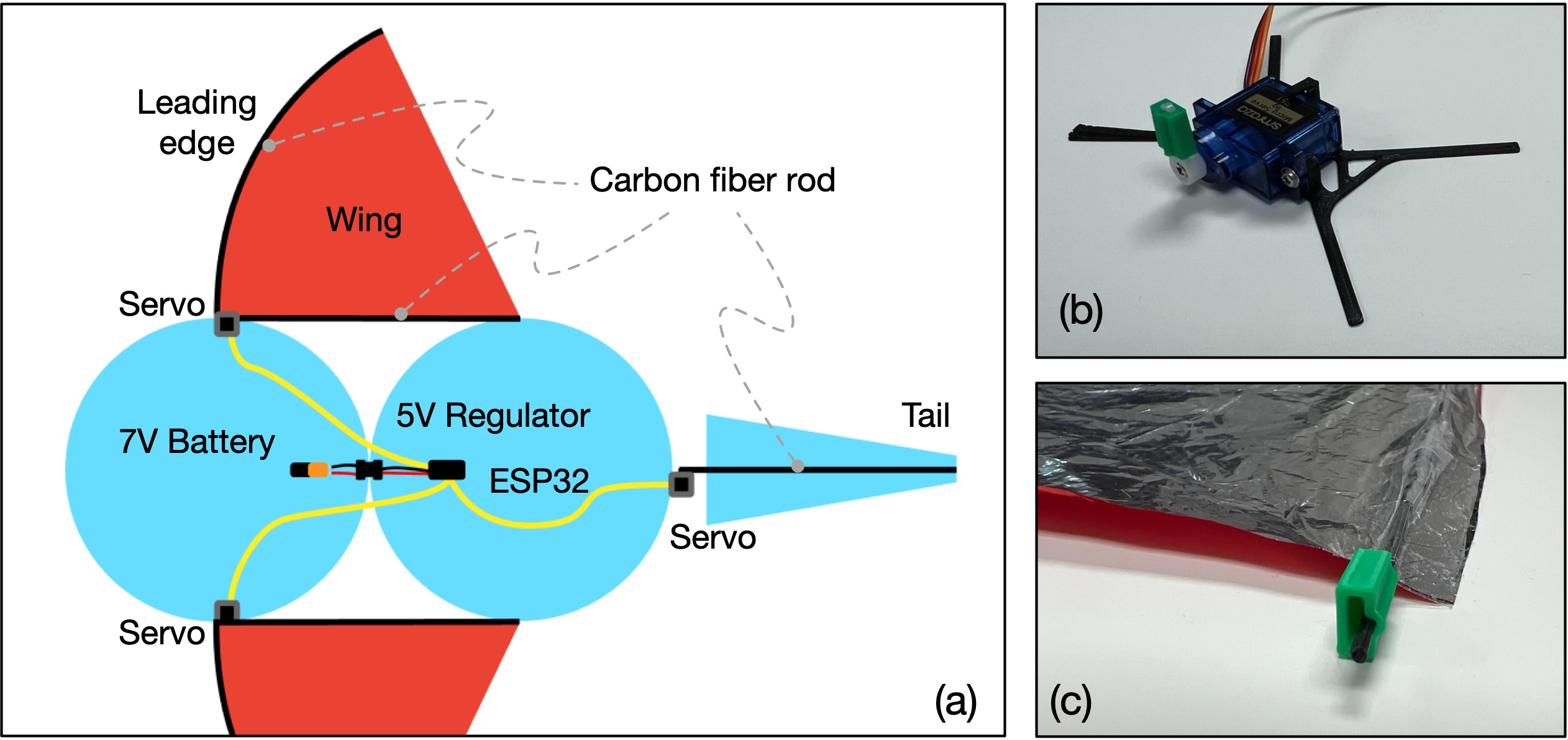}
    \caption{(a) Top-view schematic of Flappy. (b)~Servo-motor mount. (c)~Clamp for wing-servo attachment (green).}{}
    \label{fig:schematic}
\end{figure}

The wings and tail surface are made out of metalized film that is crumpled to induce wrinkles that improve flexibility. Carbon fiber rods act as the arm structure for the wings and are directly taped along the leading edge.  
For the tail, 1.5 mm carbon fiber rods are taped lengthwise through the center of a foil cutout. Depending on the shape and length, additional pieces of carbon fiber can be placed spanning the tail width to support its shape and prevent it from curling. To attach the wings and tail onto the servos, a 3D printed clamp (Fig.~\ref{fig:schematic}c) is slid over the plastic servo arm. The wings and tail all have a slight extrusion of the 1.5 mm carbon fiber rods that friction fit into the clamp. The clamp is removable by hand, allowing for modular swapping of different wing shapes or tails. Further details on wing structure and design will be discussed in Sec.~\ref{sec:flapping_wing}.


\section{Vehicle Control}
\label{sec:Dynamics}

This section describes our strategy in controlling the vehicle's velocity and attitude using the three actuators.


\subsection{Speed Control}
\label{sub:SpeedControl}

 To control the speed, the amplitude and frequency of the flaps are adjusted to change the amount of thrust generated. As will be shown later in Sec.~\ref{sec:results}, maximum thrust is achieved with the amplitude of 90$^\circ$ angle at the frequency of around 1.5~Hz. For lower speeds, the amplitude of the flapping is reduced while increasing the frequency. This is to enhance the reaction time of the speed control since low speed is typically used for precise control of the vehicle position. This speed setting is mapped to the ``throttle'' (left joystick) of the controller for intuitive control of the forward motion.

\subsection{Yaw Control}
\label{sub:YawControl}

There are two ways to control the yaw rate of Flappy: thrust differential and the tail.
For the thrust differential, the wing opposite the desired turn direction flaps, while the other wing is idle. 
The flapping amplitude is reduced to attenuate roll oscillation.
At the same time, the flapping frequency is increased to compensate for the reduced thrust and to improve the time resolution of the yaw control.

As the second approach, the tail can be used as a rudder.
When used as a rudder, the tail servo must be mounted in an orientation to allow the servo arm to rotate around the yaw axis. When turning, the tail swings in the direction of the turn. 
The aerodynamic force on the tail causes the body to turn in the yaw axis.
A byproduct of the tail turning is the slight shift forward of {center of gravity} (CG), causing Flappy to pitch down whenever turning. If turning continuously, Flappy turns similar to a ``graveyard spiral.'' 

\subsection{Pitch Control}
\label{sub:PitchControl}



Pitch control is essential for the vehicle to control its altitude.
By attaching the tail servo to rotate about the pitch axis, we can change the function of the tail from a rudder to an elevator. 
The servo swings the tail up / down to influence pitch. 
Note that this setting allows us to have both yaw and pitch control since yaw can still be achieved by thrust differential of the main wing without using the tail.

\begin{figure}[t]
    \centering
    \includegraphics[width=\linewidth]{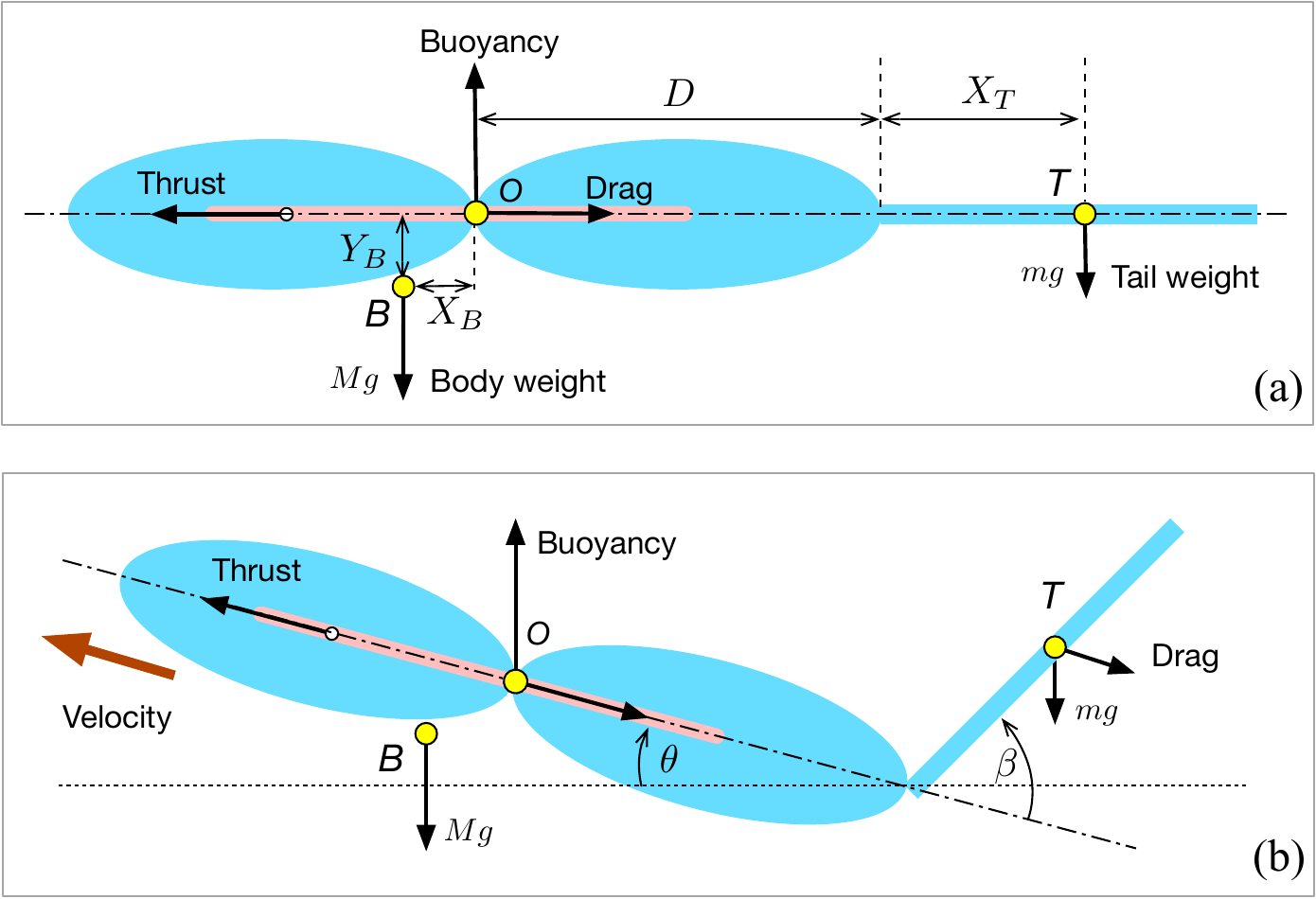}
    \caption{Longitudinal model for pitch analysis. (a) Definition of distances. (b) Definition of pitch and tail angles.}
    \label{fig:longitudinal_model}
\end{figure}

Predicting the pitch dynamics of Flappy is challenging since the tail angle affects both the longitudinal CG location and the aerodynamic force.
Here we provide a simple longitudinal model to analyze how the tail can be used to pitch the vehicle up and down. Figure~\ref{fig:longitudinal_model} depicts the simplified model used in the analysis, where $m$ describes the total mass of the tail acting on its CG at point $T$, and $M$ is the mass of the remaining part of the vehicle at $B$. 

For conciseness, we use lengths that are normalized by the diameter of the balloon: $x_B\triangleq X_B/D$, $y_B\triangleq Y_B/D$, and $x_T\triangleq X_T/D$.
We also normalize the mass by $M$ and use $\sigma \triangleq m/M$ to denote the relative mass of the tail.
Due to the symmetry, the center of buoyancy is at point $O$ where the balloons are connected. We study the moment around $O$ to identify the equilibrium pitch angle under different conditions.

We start by considering the case when the vehicle has no velocity, i.e., no aerodynamic force.
The pitch angle will be determined purely based on the balance between $m$ and $M$.
Since we want $\theta=0$ when the tail angle is $\beta=0$, the moment about the vehicle center $O$ should be balanced at this setting:
\begin{equation}
    x_B = \sigma (1+x_T).
    \label{eq:horizontal}
\end{equation}
The left-hand-side (resp.~right-hand-side) of \eqref{eq:horizontal} represents the counterclockwise (resp.~clockwise) moment about $O$, all due to the gravitational force.
The constraint \eqref{eq:horizontal} is satisfied on our vehicle by adjusting $x_B$, which is possible by changing the location of the battery and the board that are attached to the body.

\begin{figure}[t]
    \centering
    \includegraphics[width=0.93\linewidth]{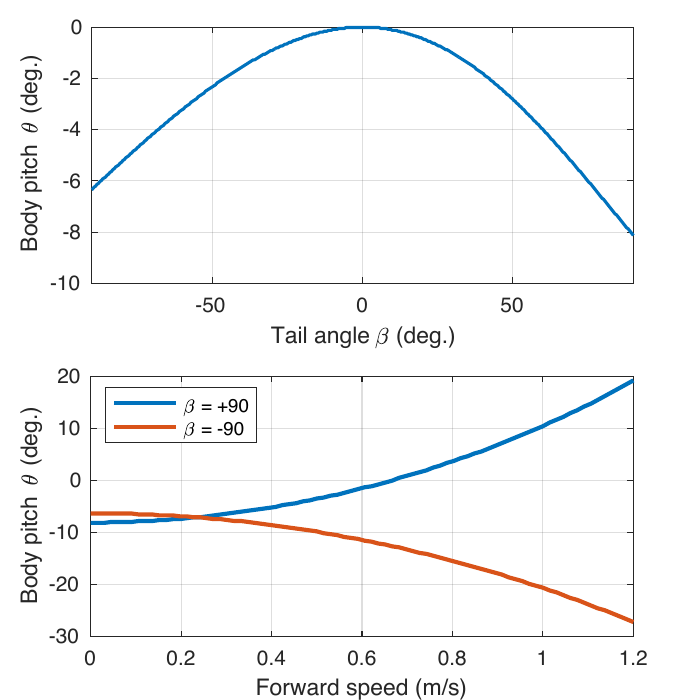}
    \caption{Analysis of pitch control. (a) The effect of tail angle on the body pitch angle when the vehicle is stationary. (b) The effect of forward velocity on the body pitch angle when the tail is $\pm90^\circ$.}
    \label{fig:pitch_plot}
\end{figure}

When the tail angle $\beta$ and the pitch angle $\theta$ are nonzero, the above moment-balance equation can be modified to the following:
\begin{equation}
    x_B \cos\theta + y_B \sin\theta = \sigma \left(\cos\theta+ x_T\cos(\beta-\theta)\right).
\end{equation}
With the current design, the vertical offset of the CG, $y_B$, is nonzero since the board and the battery are obstructed by the balloons at the center and it is convenient to attach at the bottom of the vehicle. 
With $x_B=0.075$, $y_B=0.2$, $x_T=0.5$, and $\sigma=0.05$, we have the relationship between the tail angle and the vehicle pitch as described in Fig.~\ref{fig:pitch_plot}a. 
It shows that the tail can only pitch the vehicle down due to a significant offset $y_B$. 

Now, we consider forward velocity with the assumption that it is aligned with the vehicle heading. We also approximate the tail as a rectangular flat-plate that generates drag. This model will be appropriate when the tail angle is close to $\beta = \pm 90^\circ$, and it helps us identify the maximum pitching capability of the vehicle.
The area of the tail is $A=w_T\cdot 2X_T$ where $w_T$ is the width of the tail. 
For a given forward speed $V$~m/s, the drag force is given by $F_D(V) = C_dA\rho V^2/2$ where we use the drag coefficient of $C_d=1.5$, and air density $\rho=1.225$~kg/m$^3$.

For the tail angle of $\beta=\pm 90^\circ$, the pitch angle of the vehicle as a function of the forward speed, $V$, can be obtain from the following moment balance:
\begin{equation}
    x_B \cos\theta + y_B \sin\theta = \sigma \left(\cos\theta \pm x_T\sin\theta\right) + \frac{F_D(V)}{Mg}\cdot x_T
\end{equation}
The resultant pitch angle is shown in Fig.~\ref{fig:pitch_plot}b. The figure indicates that with sufficient forward speed, the vehicle is able to both pitch up and down based on the tail position, which is verified in our experiments (see the supplementary video).
A more realistic dynamical model as well as its stability analysis are topics of future work.

\section{Flapping Wing}
\label{sec:flapping_wing}
This section describes the construction of the main wing and its control.

\subsection{Wing Structure}
\label{sub:WingStructure}

The main parameter regarding wing structure is the leading edge flexibility. We replicate this characteristic using two different carbon fiber rod diameters. The two options for the leading edge structure are Flexible or Stiff. All wings house a 1.5~mm rod at the root, which is where the leading edge is attached to the servo motor. A Stiff wing continues to use the 1.5~mm rod until the wing tip. A Flexible wing uses the bendier 1.0~mm rod after the root section to the wing tip. 

\subsection{Wing Shape}
\label{sub:WingShape}

\begin{figure}
    \centering    \includegraphics[width=0.9\linewidth]{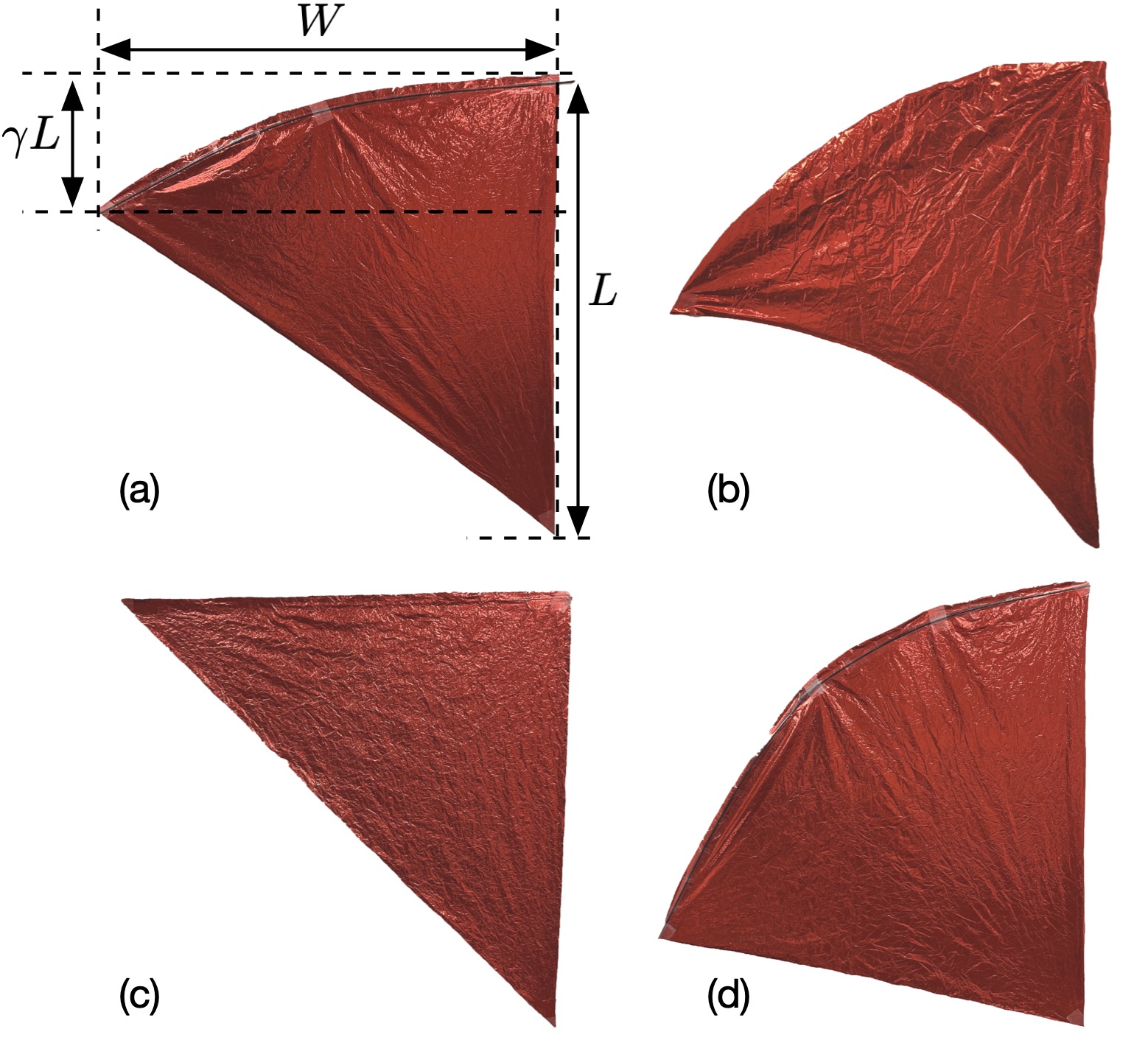}
    \caption{Wing shape and parameters. {(a) Definition of the parameters $W$, $L$, and the ratio $\gamma$. (b) Wing with concave trailing edge, $\gamma=0.5$ and $AR=1$.  (c) No sweep back, $\gamma=0$ and $AR=1$. (d) Large sweep back, $\gamma=0.75$ and $AR=1$}.}
    \label{fig:Wing Outline}
\end{figure}

We explore four parameters regarding wing shape.
Width~(\emph{W}) is the lateral distance seen in Fig.~\ref{fig:Wing Outline} of the wing arm. Length (\emph{L}) is the longitudinal distance seen in Fig.~\ref{fig:Wing Outline}.
These two parameters determine the wing size and its Aspect Ratio (\emph{AR}): the ratio of a wing's width to its length, defined as $AR\triangleq W/L$.
The third parameter is the \emph{sweep back}, which we describe using the longitudinal location of the wing tip relative to the length.
The ratio $\gamma \in[0,1]$ is used to denote the location as shown in Fig.~\ref{fig:Wing Outline}a. 
The value of $\gamma=0$ indicates a straight leading edge with no curve (see Fig.~\ref{fig:Wing Outline}c). 
When $\gamma>0$, a curvature of the leading edge is introduced according to the bending of the carbon fiber rod, since all wings attach parallel to the servo arm at the root section. The larger the $\gamma$, the greater the sweep back. 
The final parameter is the curvature of the trailing edge. We compare straight and concave trailing edges to observe their effect on thrust generation.  




\subsection{Wing Control}
\label{sub:WingControl}


The actuation of the wing servos are controlled using PWM (Pulse-Width Modulation) signals from the ESP32 Feather.
In pursuit of sinusoidal undulations in the wing, the servo actuation is controlled to follow a sinusoidal velocity profile (see Fig.~\ref{fig:Specific Wing Thrust Reading}). Because the servos only accept position inputs, and not velocity inputs, an incremental approach is used to achieve the sinusoidal profile. Servo positions are updated once every 20~ms (50~Hz), and the distance the servo is instructed to move is varied. The resulting velocity profile can be adjusted based on the desired frequency and amplitude.


\section{Testing Methods}
This section describes the thrust measuring device we use to explore the parameter space of wing shapes and structure. We also describe how we compare our flapping-wing vehicle with a more conventional propeller-based design in terms of energy efficiency.

\subsection{Thrust Measurement}
A thrust measuring jig is constructed to analyze the effect of wing parameters. The concept is similar to propeller testing stands where a rotating arm pushes down on a scale as thrust is produced \cite{PropStand}. Unlike propellers, flapping wings do not produce a constant thrust value on a single point. To account for the oscillatory motion of flapping wings, we need scales at the front and back of the wing to capture the entire wing's thrust generation. 

\begin{figure}
    \centering    \includegraphics[width=\linewidth]{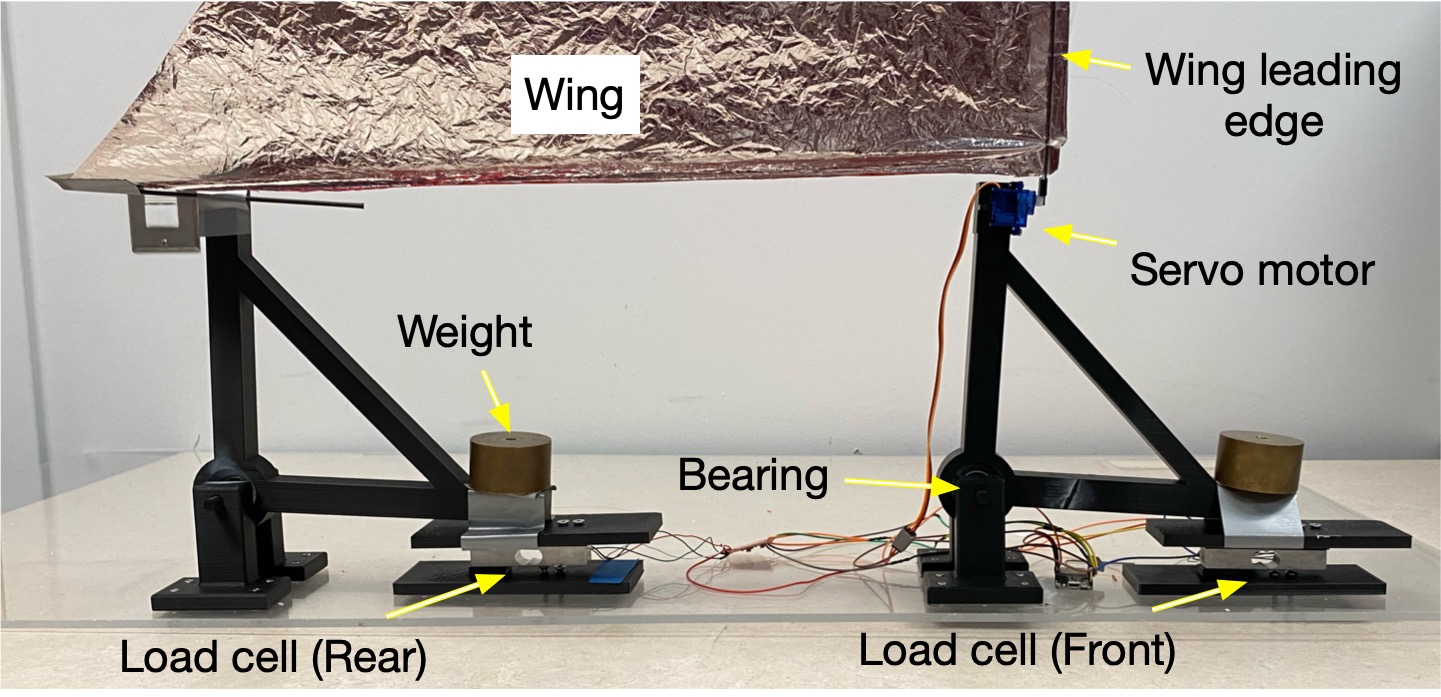}
    \caption{Thrust tester with wing setup.}
    \label{fig:Wing Thruster}
\end{figure}

A right-angle arm is supported with ball bearings around a shaft held by stands. A 1~kg load cell is secured between two plates. The horizontal arm is placed on top of the load cell plates. 
The same 9~g servo we use on Flappy is attached to the top of the front arm. (The arm can be extended to enhance the resolution of thrust measurement.) The servo is connected to an ESP32 Feather. Wings are secured onto the servo using the same method they attach to Flappy. 
Each load cell is wired to an HX711 amplifier, which are both connected to the ESP32 Feather. The load cells are calibrated to measure thrust in grams. Figure \ref{fig:Wing Thruster} shows the complete set up with a wing attached.

Each arm has only one degree of motion rotating around its bearing pivot point. As the wing flaps, only forward thrust will cause the arm to generate a thrust reading on the load cells. The weights hold the arm down onto the load cell and allow us to measure negative thrusts during the flap upstroke. Prior to each test run, the load cells will tare with the wing held horizontally. 

Flapping amplitude and frequency affect thrust generation~\cite{yu2014lift}. For each wing, we run a cycle of tests covering a range of amplitudes and frequencies. We calculate the average thrust generated over a set time and find the optimal flap settings that generate the greatest average thrust. These tests are repeated for every wing parameter combination 
to study the effect of
wing shape and structure.   

\subsection{Energy Consumption}

Performance of Flappy will be compared against a propeller-based agent, "Proppy", to see the efficiency of flapping wings propulsion. Proppy is a replica of Flappy with two propeller motors replacing the wings. Several different motors and propellers are tested to determine the most optimal set to maximize Proppy's range. The motors chosen are compatible with our electronics, within the agent's payload budget, and of similar weight class to our flapping wing system. All other components remain the same, including the body shape and tail. The three performance parameters to compare are endurance, speed, and maximum range. 

Endurance is the maximum duration of time our agent is operable given a power supply. Using the same 2s 300~mAh battery for both Flappy and Proppy, we can compare the rate of energy consumption. From a fully charged battery, we set both agents to fly forward continuously and measure the time until the battery is drained and the agent stops. Several trials are run at various speed control settings to test if energy consumption is affected at higher speeds.

Range is the maximum distance an agent can travel given a power supply. With both endurance and speed data, the maximum range of the agent at certain speed settings can be calculated by $Range = Endurance \times Speed$.  We will compare the maximum possible range of the agents and observe if speed and endurance have an adverse relation. 


\section{Experimental Results}
\label{sec:results}
\subsection{Effect of Wing Structure and Shape}


To identify the effect of wing parameters, we tested wings with varying shapes and structures. In terms of shapes, all the wing lengths were constrained as the distance between the two load cell arms at 53~cm. The \emph{AR} values chosen were 0.70, 0.85, 1.0, and 1.15, which converts to widths of 37, 45, 53, and 61~cm. Four $\gamma$ values were tested at 0.0, 0.25, 0.5, and 0.75. Both straight and concave trailing edges were tested for each combination of the shape parameters.

\begin{figure}[b]
    \centering
    \includegraphics[width=\linewidth]{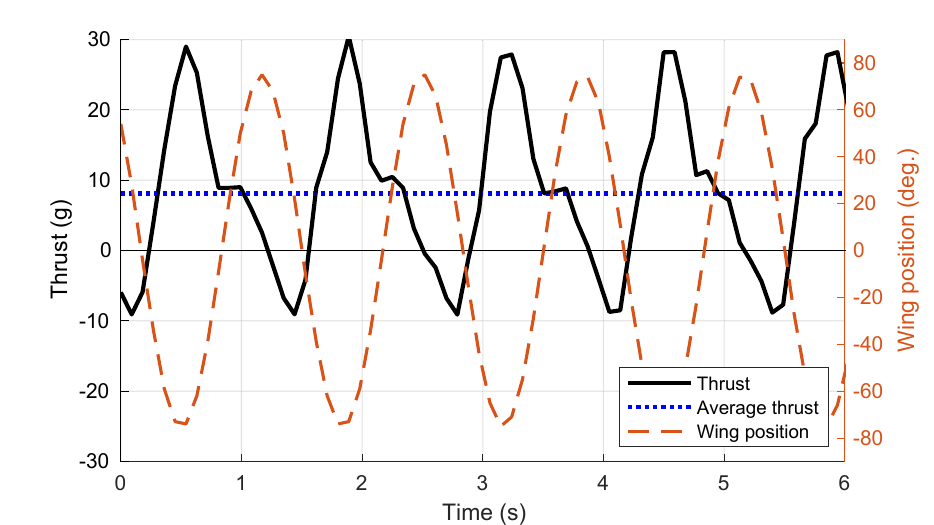}
    \caption{Time history of thrust generated from a base wing using flapping amplitude of $75^\circ$ and frequency of 0.75~Hz.}
    \label{fig:Specific Wing Thrust Reading}
\end{figure}

For a given wing, a selected amplitude and frequency pair generates a time history of the thrust produced as shown in Fig.~\ref{fig:Specific Wing Thrust Reading}.
The wing position is denoted by the angle with respect to the ground plane: $+90$ corresponds to the highest position.
We use the time average to capture the net forward force that is generated by the wing. Figure~\ref{fig:Specific Wing Thrust Reading} also shows that small negative thrusts are produced after the completion of the wing's upstroke, and the peak thrust is achieved after the completion of the downstroke. 

\begin{figure}[t]
    \centering
    \includegraphics[width=\linewidth]{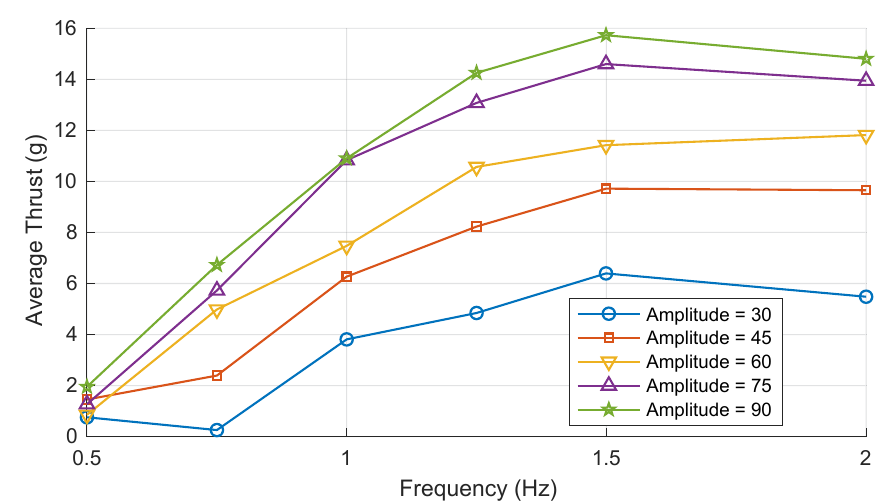}
    \caption{Average thrust at different amplitudes and frequencies (for $W=53$~cm, $AR=1.0$, and $\gamma=0.5$).}
    \label{fig:Average Wing Thrust Reading}
\end{figure}

The average thrust for every pair of amplitude and frequencies are plotted in Fig.~\ref{fig:Average Wing Thrust Reading}. Each line represents a different amplitude setting across different frequencies. The frequency was limited to 2 Hz as the servo cannot move fast enough to achieve full motion when paired with high amplitude values. Testing different servos that have a higher operating speed would resolve this issue, but such servos are typically heavier and we constrain our test to use same servos viable for Flappy. For each wing shape, the thrust value attained at the optimal flap setting is assigned.
In the parameter regime that we have tested, the optimal amplitude was either of the highest setting, 75$^\circ$ or 90$^\circ$, and the optimal frequency was typically around 1.0 to 1.5~Hz.
With the maximum average thrust value of each wing, we study the effect each parameter has on the thrust.

\subsubsection*{Sweep Back}
\begin{figure}[b]
    \centering
    \includegraphics[width=\linewidth]{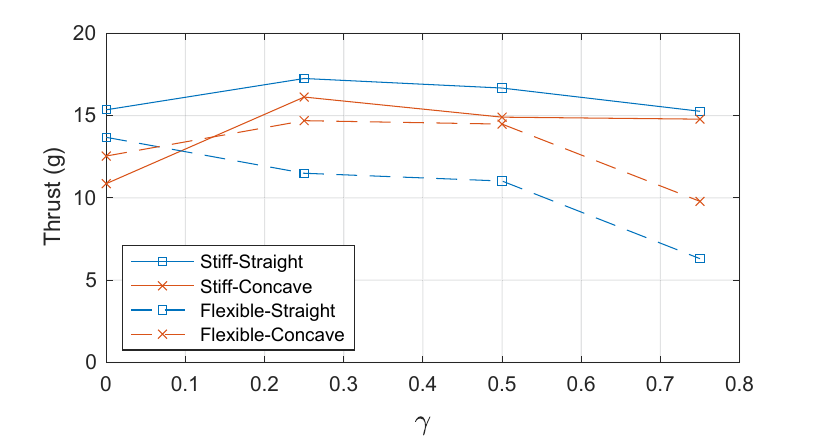}
    \caption{The effect of sweep back parameter $\gamma$ on thrust. Width of 53~cm and \emph{AR} of 1.0 is used.}
    \label{fig:WTvsThrust}
\end{figure}

To see if sweeping the leading edge offered any benefit, the thrust readings were compared to a wing without a leading edge curve. 
Figure~\ref{fig:WTvsThrust} compares the average thrust values of a given wing with fixed \emph{AR} and varying $\gamma$.
At $\gamma=0$, the wing is shaped like a right triangle. 
From Fig.~\ref{fig:WTvsThrust}, stiff wings seem to slightly improve the thrust capability with sweep back. However, the Stiff straight wing did not seem to benefit from sweep back.
Furthermore, greater sweep backs ($\gamma=0.75$) tend to decrease the thrust especially for the Flexible wings.

\subsubsection*{Aspect Ratio}
\begin{figure}[h]
    \centering
    \includegraphics[width=\linewidth]{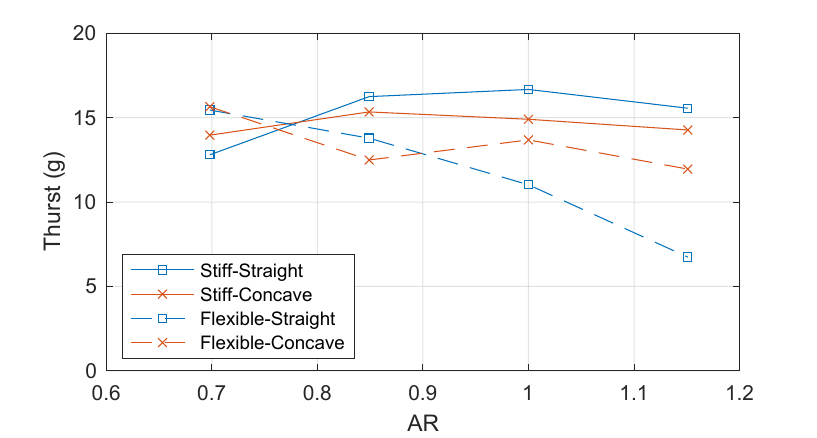}
    \caption{The effect of \emph{AR} on thrust for $L=53$~cm and $\gamma=0.5$.}
    \label{fig:ARvsThrust}
\end{figure}

The effect of \emph{AR} is tested with wings with fixed $\gamma=0.5$ and $L=53$~cm. As seen in Fig.~\ref{fig:ARvsThrust}, the Flexible wings tend to produce less thrust with higher \emph{AR} (i.e., larger $W$ value).
Interestingly, Flexible wings prefer and outperform the Stiff wings at the smallest \emph{AR} (smaller \emph{W}) but suffer as the wing gets wider. This degradation at high \emph{AR} may be caused by the degradation in the flapping motion since the force translated to the wing tip is insufficient. 
The Stiff wings seem to be less sensitive to \emph{AR}. 
{Notably, the best frequency setting was highly sensitive to \emph{AR}, or the wing width (not shown in this figure). It ranged from 0.75~Hz for $W=61$~cm to 1.50~Hz for $W=37$~cm.}

\begin{table*}[b]
\centering
    \begin{tabular}{ |c|c|c c|c c|c c|  }
        \hline
        &
        &
        \multicolumn{2}{c|}{T4030 Bi-Blade}&
        \multicolumn{2}{c|}{T3020 Bi-Blade}&
        \multicolumn{2}{c|}{2525 Tri-Blade}\\
        \hline
        Motor& Duty Cycle \%& Endurance (s)& Thrust (g)& Endurance (s)& Thrust (g)& Endurance (s)& Thrust (g)\\
        \hline
        \multirow{5}{4em}{XING2 1404 3800KV}& 100& 145& 145& 269& 85.5& \cellcolor[HTML]{EC7063} 188& \cellcolor[HTML]{EC7063}51\\
        &80& 216& 113& 369& 65& \cellcolor[HTML]{EC7063}288& \cellcolor[HTML]{EC7063}39.8\\
        &60& 388& 83.5& 661& 41.5& \cellcolor[HTML]{EC7063}502& \cellcolor[HTML]{EC7063}26.3\\
        &40& 757& 48& 1294& 19.4& \cellcolor[HTML]{EC7063}1060& \cellcolor[HTML]{EC7063}14\\
        &20& 2363& 15.5& 2983& 4.85& \cellcolor[HTML]{EC7063}2745& \cellcolor[HTML]{EC7063}4.12\\
        \hline
        \multirow{5}{4em}{XING 1404 4600KV}& 100& 111& 145& \cellcolor[HTML]{EC7063}158& \cellcolor[HTML]{EC7063}100& 162& 112\\
        &80& 160& 125& \cellcolor[HTML]{EC7063}238& \cellcolor[HTML]{EC7063}79.5& 246& 92\\
        &60& 244& 90& \cellcolor[HTML]{EC7063}387& \cellcolor[HTML]{EC7063}52& 385& 63\\
        &40& 434& 55.2& \cellcolor[HTML]{EC7063}668& \cellcolor[HTML]{EC7063}27& 743& 35.5\\
        &20& 1448& 19& \cellcolor[HTML]{EC7063}1766& \cellcolor[HTML]{EC7063}6.14& 2261& 11.45\\
        \hline
        \multirow{5}{4em}{XING Nano 1303 5000KV}& 100& \cellcolor[HTML]{EC7063}138& \cellcolor[HTML]{EC7063}121& 217& 100& \cellcolor[HTML]{EC7063}155& \cellcolor[HTML]{EC7063}90\\
        &80& \cellcolor[HTML]{EC7063}199& \cellcolor[HTML]{EC7063}100& 301& 75& \cellcolor[HTML]{EC7063}230& \cellcolor[HTML]{EC7063}75\\
        &60& \cellcolor[HTML]{EC7063}324& \cellcolor[HTML]{EC7063}75& 494& 50.5& \cellcolor[HTML]{EC7063}377& \cellcolor[HTML]{EC7063}52\\
        &40& \cellcolor[HTML]{EC7063}657& \cellcolor[HTML]{EC7063}45& 1051& 25.8& \cellcolor[HTML]{EC7063}756& \cellcolor[HTML]{EC7063}25.5\\
        &20& \cellcolor[HTML]{EC7063}1876& \cellcolor[HTML]{EC7063}16& 2576& 7.1& \cellcolor[HTML]{EC7063}2188& \cellcolor[HTML]{EC7063}9.1\\
        \hline
        
    \end{tabular}
\caption{Endurance and thrust readings from each motor and propeller blade pair. The red pairs were omitted from speed testing as both their endurance and thrust readings were inferior to similar pairs. The rest were speed tested with Proppy in Figure \ref{fig:Flappy vs Proppy}.}
\label{table:1}
\end{table*}

\subsubsection*{Trailing Edge}
The trailing edge parameters are represented in both  Fig.~\ref{fig:WTvsThrust} and Fig.~\ref{fig:ARvsThrust}. 
With the wings we tested, we cannot draw definitive conclusion regarding the effect of the trailing edge shape. 

\subsubsection*{Leading Edge Stiffness}
We performed tests for two stiffness settings.
Note that even though we refer to them as Stiff wings, the 1.5~mm rods are not completely rigid and do offer flexibility. 
For the majority of wings with only the leading edge stiffness changed, the stiffer wings produced a greater average thrust. Besides rare cases like the aforementioned advantage the Flexible wings have at the smallest \emph{AR} and without sweep back, the best thrust options are with the Stiff wings as those parameters increase. Testing with higher stiffness settings is subject of our ongoing work. 

\subsubsection*{Optimal Wing}
From our results, the optimal wing is a Stiff $AR=1.0$, $\gamma=0.25$, wing with a concave trailing edge. With 90$^\circ$ at 1.25~Hz, the wing generates an average thrust of 17.3~g.




\subsection{Energy Efficiency}

\begin{figure}[t]
    \centering
    \includegraphics[width=\linewidth]{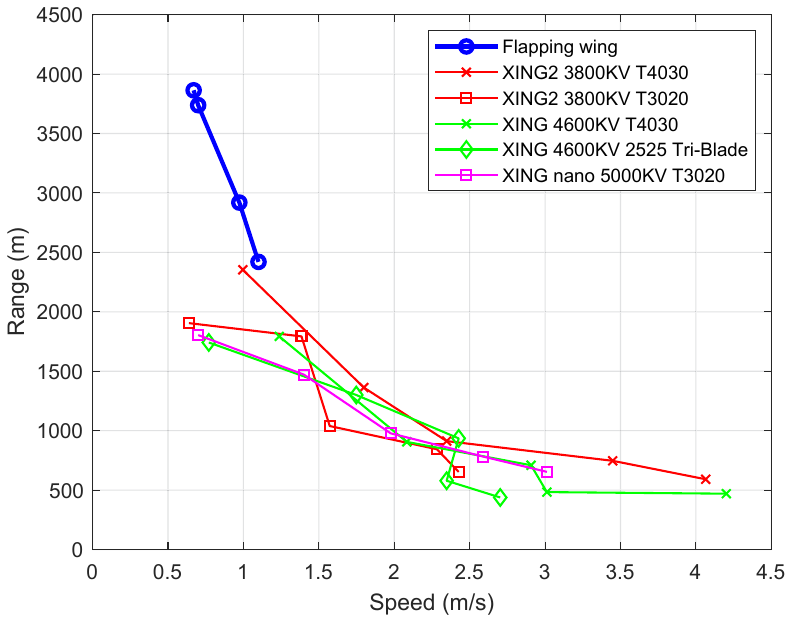}
    \caption{Comparison between propeller-based and flapping-based designs in terms of speed and range.}
    \label{fig:Flappy vs Proppy}
    
\end{figure}

With continuous flaps, endurance of Flappy was not affected by the speed settings on the wings. No matter the frequency or amplitude of the flaps, the endurance of Flappy with a fully charged 2s 300~mAh LiPo battery was on average 2200 seconds, around 37 minutes. By equipping Flappy with the optimal wings, we measured a maximum speed of 1.1~m/s. For continuous flaps, maximum range of Flappy is calculated to be 2420~m. If intermittent pauses between flaps are introduced, similar to how gliding birds increase their flight range \cite{harvey2021aerodynamic}, the same concept applies to Flappy to increase range. Flappy stays powered on for 2 hours and 40 minutes when none of the servos actuate. Taking this into account, we speed tested Flappy with intermittent flaps. As seen in Figure \ref{fig:Flappy vs Proppy}, Flappy's average speed is decreased but range is substantially increased.

Three different motors and propellers were picked to find the most optimal pairing. The endurance of each motor and propeller pair was tested, as different propellers cause different loads on the motors. We set different duty cycles percentages of the PWM signal, as this controls the amount of power being delivered to the ESC. Higher duty cycle \% of the motor leads to higher rotational speed (rpm) and higher thrust, but at the cost of endurance. Thrust readings were collected using a similar testing jig as the flapping-wings. From the test results in Table \ref{table:1} and Figure \ref{fig:Flappy vs Proppy}, the motor XING2 1404 3800KV and the propeller blade Nazgul T4030 Bi-Blade was chosen as the optimal pair.

Figure \ref{fig:Flappy vs Proppy} shows the range of the agents at various speeds. At the maximum speed setting, Proppy moves exceptionally faster than Flappy, but at the cost of reduced endurance of only 162 seconds. The maximum range setting of Proppy was with the duty cycle set to 20\%.
Unlike the flapping-wing system, the thrust generated by propeller motors decrease as the battery capacity drains, slowing the agent down. To account for this, we tested Proppy's speed with batteries of different charges at this optimal 20\% duty cycle. These batteries were from running Proppy for 10, 20, and 30 minutes, which resulted in batteries of charge 70\%, 40\%, and 21\% respectively. The speed of Proppy at these charges were 4.8~m/s, 5.2~m/s, and 5.6~m/s respectively. By taking the average including the full charge speed, the speed of Proppy to maximize range is 1~m/s. The endurance improves to 2350~seconds, around 39 minutes, increasing the maximum range for Proppy to 2350~m.

Comparing the ranges at around 1.1~m/s, Flappy beats Proppy's maximum range for all propeller motor configurations. Proppy's maximum range may be greater at even lower speed settings, but the motors are unable to effectively turn at such low speeds. Additionally, the propeller motor system is 20~g heavier than the flapping-wing system. This weight difference allows for additional components to be added to Flappy. Lastly, the propeller motors are dangerous when operated, and risk of injury to other agents or people is great when in close proximity. For similar size LTA vehicles that only need to operate at speeds from 0.5~m/s to 1~m/s, flapping wings are the viable and safer option to last longer and travel farther.

\section{Discussion}
\paragraph*{Improving experimental data} Obtaining a conclusive pattern from the thrust measurements was challenging due to the variance in the plot. We believe that this comes from the variance in the fabrication quality of the wings. The experimental results can be improved by increasing the number of wings tested per data point: i.e., test multiple wings with the same shape parameters. Improvement in the design to help with consistency in the fabrication quality is also an interesting avenue for future work.

Additionally, a motion capture system to accurately measure dynamics and performance metrics of the agent's would allow better comparative and predictive data. Instead of obtaining speed and range through calculation, a log of data through the agent's operation would be more desirable. 


\paragraph*{Modularity}
Our vehicle has high modularity in terms of how to connect the balloons and where to place the wings.
We are investigating the possibility of adding sensors and autonomy to this vehicle, which can be done by increasing the number of balloons for higher payload. Miniaturization is also possible by working with one balloon with smaller wings and tails.

\paragraph*{Autonomous Controls Problem}
The next step for Flappy is to fly autonomously. An applicable example is to add a camera in the front to detect a target and pursue. For a standard propeller blimp the control system takes advantage of the constant thrust from the propeller and tunes the gain accordingly. This option is not as intuitive for flapping-wings, due to their oscillatory thrust generation.

\paragraph*{Simplicity}
Flappy is also very easy to make with components available off-the-shelf (e.g., Amazon), with the exception of helium. Thanks to this, Flappy has been showcased at various demos and STEM workshops, including IROS 2023, AISES 2023 Conference, STEM Santa Fe, and more. Teaching and building Flappy is a great pathway to core concepts of mechatronics, programming, circuits, aerodynamics, and CAD modeling. Students are free to experiment and come up with new designs using the Flappy platform. 

\section{Conclusion}
We present a design of a flapping-wing LTA vehicle whose wings are inspired by manta rays. The vehicle achieves yaw control with thrust differential from two wings and pitch control using the tail.
We performed parametric analysis of the wing and identified an optimal combination of shape, structure, and flapping setting for the 9~g servo motor that is widely available commercially.
Our experimental result suggests that our flapping-wing LTA vehicle achieves higher efficiency when compared to a rotational propeller-based counterpart, as the range is greater than comparable propeller motors.
The direction of future work includes a more detailed fluid-mechanical analysis of the wings, the development of miniaturized version of the vehicle, improvement in the flapping with linkage mechanism, and the implementation of autonomy.




\bibliographystyle{ieeetr}
\bibliography{flappy}

\end{document}